\documentclass[a4paper,twocolumn]{article}
\usepackage{eurosis}
\usepackage{graphicx}
\usepackage{myHarvard}
\usepackage{url}
\usepackage[utf8]{inputenc}

\title{AUTOMATABLE EVALUATION METHOD\\ORIENTED TOWARD BEHAVIOUR BELIEVABILITY\\FOR VIDEO GAMES}
\author{
Fabien Tenc\'e$^{*,**}$ and C\'edric Buche$^*$\\
$^*$ Universit\'e Europ\'eenne de Bretagne - ENIB\\
LISyC - CERV\\
25, rue Claude Chappe\\
29280 Plouzan\'e (France)\\
$^{**}$ Virtualys\\
41 rue Yves Collet\\
29200 Brest (France)\\
\{tence,buche\}@enib.fr
}
\date{}

\begin{document}

\maketitle
\thispagestyle{empty}


\begin{abstract}
Classic evaluation methods of believable agents are time-consuming because they involve many human to judge agents. They are well suited to validate work on new believable behaviours models. However, during the implementation, numerous experiments can help to improve agents' believability. We propose a method which aim at assessing how much an agent's behaviour looks like humans' behaviours. By representing behaviours with vectors, we can store data computed for humans and then evaluate as many agents as needed without further need of humans. We present a test experiment which shows that even a simple evaluation following our method can reveal differences between quite believable agents and humans. This method seems promising although, as shown in our experiment, results' analysis can be difficult. 
\end{abstract}

\section{INTRODUCTION}
\paragraph{}
The vast majority of video games features computer controlled virtual characters, also known as agents, to enrich their environment. For games to be able to suspend disbelief \cite{Bates1992}, those agents should be \emph{believable}. However, this term have two distinct meanings: first, the ability to be a good role-player and second, the ability to be indistinguishable from a human player \cite{Livingstone2006}. We aim at agents taking on the role of players, so we will use the latter definition in this paper.
\paragraph{}
When modelling or implementing an agent's behaviour, an important step is evaluation. It can be very useful to improve the final result by tuning the model's parameters or by modifying the model itself. Evaluation is almost mandatory to validate work and should be able to answer: to what extent does the agent meet the objectives? or is it better than what have been done before? However, evaluation of believable agents is rather difficult because believability is based on the observers' feelings. Many studies deal with this problem like \cite{MacNamee2004,Livingstone2006,Gorman2006}. However, they still rely on humans to judges agents. As a result, this kind of method cannot be used to evaluate a large number of different behaviours. Moreover, the use of questionnaires for evaluation is criticised \cite{Slater2004}.
\paragraph{}
Knowing these problems, we propose a method using a different approach: we aim at reducing human intervention as much as possible still helping researchers assessing artificial behaviours' believability. The objective is to have a  method which could be automatised and thus which can be used in optimisation algorithms or for tests during the night. For this, we chose to measure artificial behaviours' likeness to human behaviours, a concept close to believability. Of course, this method does not aim at replacing classic ones but rather offers a complementary approach so that the final result looks more believable.

\section{EVALUATION METHOD PROPOSITION}
\subsection{Principle}
\paragraph{}
The evaluation's general principle is to compute vectors which are features of humans' and agents' behaviours. We call those vectors ``signatures'' as they should be designed to be representative of a behaviour. By measuring the distance between an agent's signature and humans', it should be possible to tell if its behaviour looks like a human behaviour or not.
\subsection{Protocol}
\paragraph{}
Here are the main stages of the protocol to evaluate an agent with respect to humans:
\begin{enumerate}
 \item Define behaviour signatures and metrics between them;
 \item Monitor humans (the more the better) and compute the signatures;
 \item Monitor the agent to be evaluated and compute the signatures in the same conditions as the monitored humans;
 \item Compute the distance between each agent's signature and its human equivalent.
\end{enumerate}
\paragraph{}
The first part is very challenging because we need to find signatures having a low variance for humans, still being able to detect non-believable behaviours. This could be a problem when humans are given a great freedom because their behaviour can vary widely. To determine good signatures, it can be necessary to run the protocol several times with few humans and to modify signatures and/or metrics so that they have nice properties.
\paragraph{}
For the second and the third steps, human and agents must be studied under the same conditions. Since agents do not play against humans, if virtual characters coexist in the same environment we have to make a concession: for the step 2, humans are together and for the step 3, agents are together. If we introduce agents in step 2, signatures could be flawed by reflecting interaction between humans and non-believable agents.
\paragraph{}
We presented our method as been automatable. Indeed, step 3 and 4 can be redone without further need of humans. However, automatable does not mean fast as step 3 can take some time. To determine the experiment's duration, a study should be done in order to test how much time the signatures need to stabilise.
\subsection{Monitoring agents and humans}
\paragraph{}
As explained in the protocol, we must monitor both humans and agents so that we can compute signatures. To be totally independent from the internal architecture, we chose to take the same point of view as a judge looking over the subject's shoulder. Therefore, in the well-known perception-decision-action loop we can only have access to the actions and the perceptions. We assume that the loop can be observed for human as well so that we can compare agents' and humans' perceptions and actions in the same manner.
\paragraph{}
The principle of the monitoring is to keep track of a subset of these perceptions and actions during simulations so as to build the signatures. For example, in a first person shooter, we could track basic actions and perceptions such as if the actor is jumping and if it sees an enemy. It could then be possible to build a very simple signature measuring the proportion of jumps when an enemy is visible.
\paragraph{}
There are three main kind of signatures: perception-based, action-based and those linking actions to perceptions. Perception-based signatures are not very useful because players judges agents on their actions. Action-based can be useful but may be too simple to explain complex behaviours. Finally, signatures linking actions to perceptions are the most interesting ones because they may find patterns in the decisions. However, there is a difficulty: if an information is monitored in perceptions, it is not sure that the agent or the human noticed this information.

\section{EXPERIMENT}
\begin{figure}[!ht]
	\centering
	\includegraphics[width=0.8\linewidth]{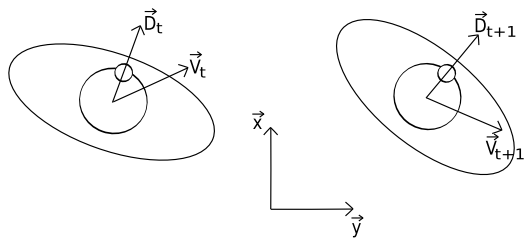}
	\caption{Scheme of a virtual character viewed from top view at two following time steps, $t$ and $t+1$. $\protect\overrightarrow{D}$ is the direction vector, pointing toward the character's aiming point and $\protect\overrightarrow{V}$ is the velocity vector.}
	\label{vectors}
\end{figure}

\begin{figure}[!ht]
	\centering
	\includegraphics[width=0.9\linewidth]{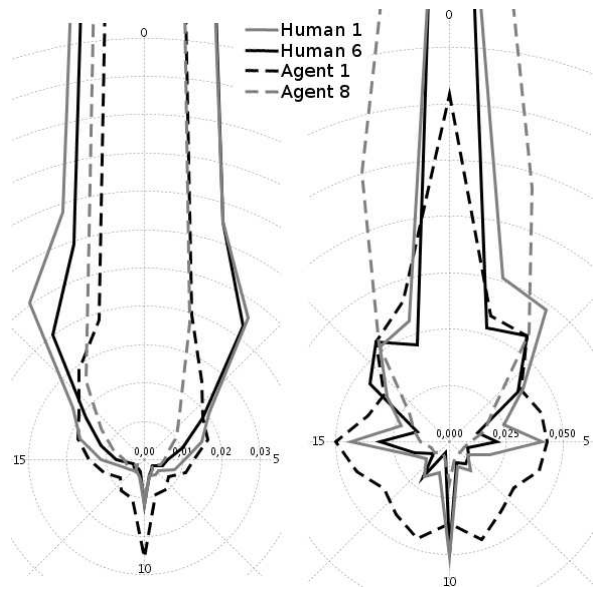}
	\caption{Sample of the signatures. On the left ``velocity change angle'' signatures and on the right  ``velocity relative to the direction'' signatures. The graphs are truncated for the $0^{th}$ component for the sake of readability.}
	\label{signatures}
\end{figure}

\paragraph{}
For the experiment, we used the game Unreal Tournament 2004. It is a first person shooter game, in other words each player or agent controls a unique virtual character and sees through its eyes. The character can, non-thoroughly , grab items (weapons, \dots{}), move (walk, jump, \dots{}) and shoot with a weapon. Each character have an amount of hit points, also known as life points: each time an actor is hit by an enemy fire, a certain amount of hit points are subtracted to the current value. When hit points reaches zero, the character ``dies'' but can usually reappear at another place in the virtual environment. Although the concept can seem very basic, it can prove challenging for agents to mimic humans.
\paragraph{}
This experiment's main objective is to give a concrete example and to show the interest of the method. Therefore, the signatures are simple and would be insufficient for a real evaluation. Moreover, for the sake of simplicity, this sample experiment try to assess only the believability of Unreal Tournament 2004 agents' movements. Here are the signatures:
\begin{itemize}
 \item Velocity change angle: 20-dimensions vector, the value of the $i$-th component is the number of time we measured an angle of approximatively $i$ between $\overrightarrow{V}_t$ and $\overrightarrow{V}_{t+1}$. Note that the vector as only 20 components so an angle of 20 corresponds to a whole turn, 10, half turn, etc. This vector is then normalised.
 \item Velocity relative to the direction: 20-dimensions vector, the value of the $i$-th component is the number of time we measured an angle of approximatively $i$ between $\overrightarrow{V}_t$ and $\overrightarrow{D}_t$. The same unit as the previous signature is used for angles. This vector is then normalised.
\end{itemize}
where $t$ is the current time step and $t+1$ is the following one, time steps occurring every 125 ms. $\overrightarrow{V}$ is the velocity vector without its $z$ component, $\overrightarrow{D}$ is a vector pointing toward the character's aiming point without its $z$ component (figure \ref{vectors}). The $z$ component was ignored for the signatures because it simplifies the signatures without losing too much information.
\paragraph{}
The experiment is composed of two steps:
\begin{enumerate}
 \item 6 low/medium level human played a \emph{deathmatch} game during 20 minutes during which signatures were computed. We chose a duration of 20 minutes because signatures tend to stabilise after 15 to 20 minutes. A \emph{deathmatch} follows the basic rules and the objective is to ``kill'' as many other players as possible. Each human plays using a keyboard and a mouse to control the virtual character and a classic desktop computer screen to see the virtual environment.
 \item Signatures are computed for 8 agents having 8 different skill levels. Those agents fight against each other in 6-agent matches in the same topography, following the same rules and during the exact same time.
\end{enumerate}
We used tools developed by our team and based on Pogamut 2 \cite{Burkert2007}. They are available at \url{svn://artemis.ms.mff.cuni.cz/pogamut/branches/fabien_tence}. The experiment gave us 2 signatures for each of the 14 subjects, each signature having 20 components. A sample of those results is given in figure \ref{signatures}.
\section{RESULTS' ANALYSIS}
\begin{figure}[!ht]
	\includegraphics[width=\linewidth]{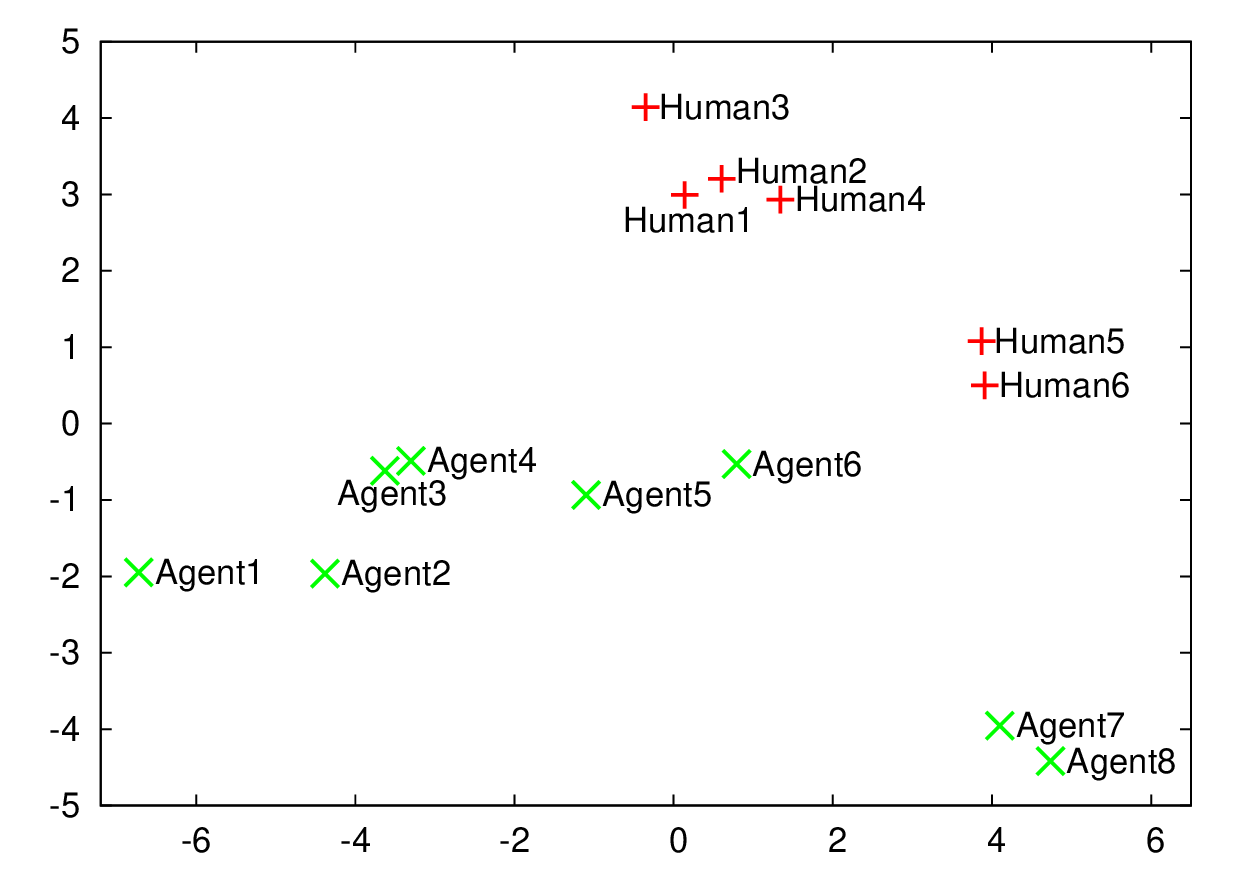}
	\caption{Subjects represented using the two principal components which explain 90.37\% of the variance for the ``velocity change angle'' signatures.}
	\label{PCAv}
\end{figure}

\begin{figure}[!ht]
	\includegraphics[width=\linewidth]{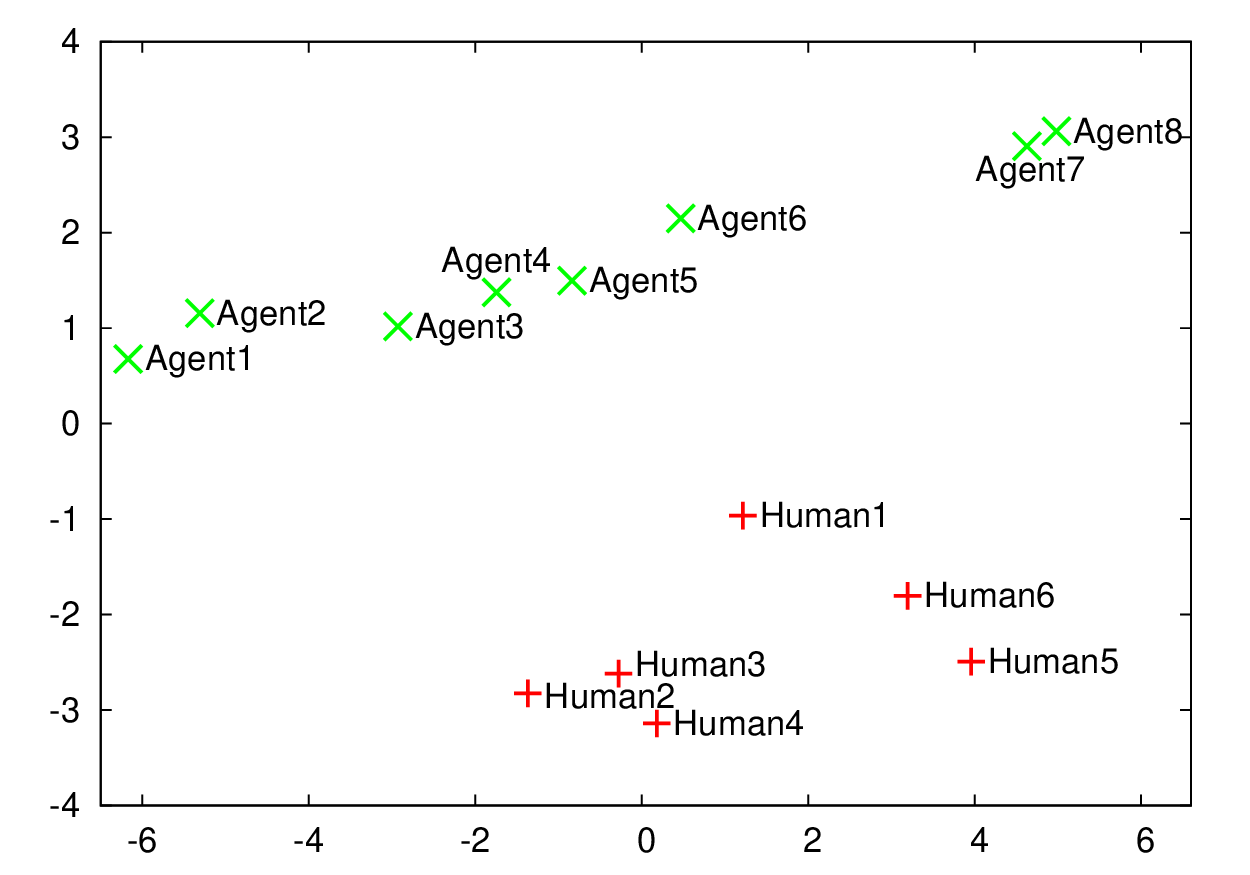}
	\caption{Subjects represented using the two principal components which explain 78.18\% of the variance for the ``velocity relative to the direction'' signatures.}
	\label{PCAdv}
\end{figure}

\begin{figure}[ht]
\begin{minipage}[c]{0.57\linewidth}
\centering
\includegraphics[width=\linewidth]{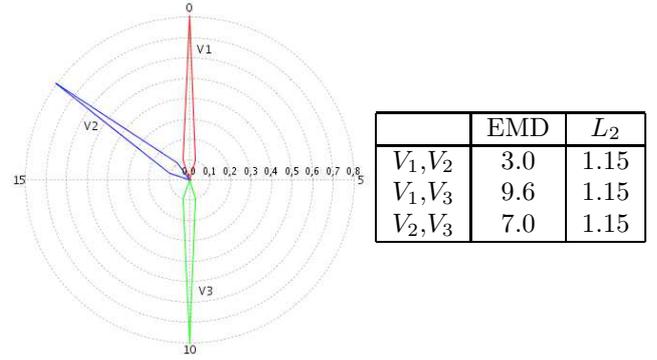}
\end{minipage}
\hspace{0.5cm}
\begin{minipage}[c]{0.35\linewidth}
\centering
	\hspace{-0.8cm}
	\begin{tabular}{|l|c|c|}
		\hline
		&EMD&$L_2$\\
		\hline
		$V_1$,$V_2$&3.0&1.15\\
		$V_1$,$V_3$&9.6&1.15\\
		$V_2$,$V_3$&7.0&1.15\\
		\hline
	\end{tabular}
\end{minipage}
\caption{Example showing the interest of the EMD compared to Euclidean distance ($L_2$) in our case. The normalised vectors not having any common component, the $L_2$ distance is then equal between each other. However, it is obvious that $V_1$ and $V_2$ are the most similar vectors which is detected by the EMD.}
\label{EMD}
\end{figure}

\begin{figure}[!ht]
	\includegraphics[width=\linewidth]{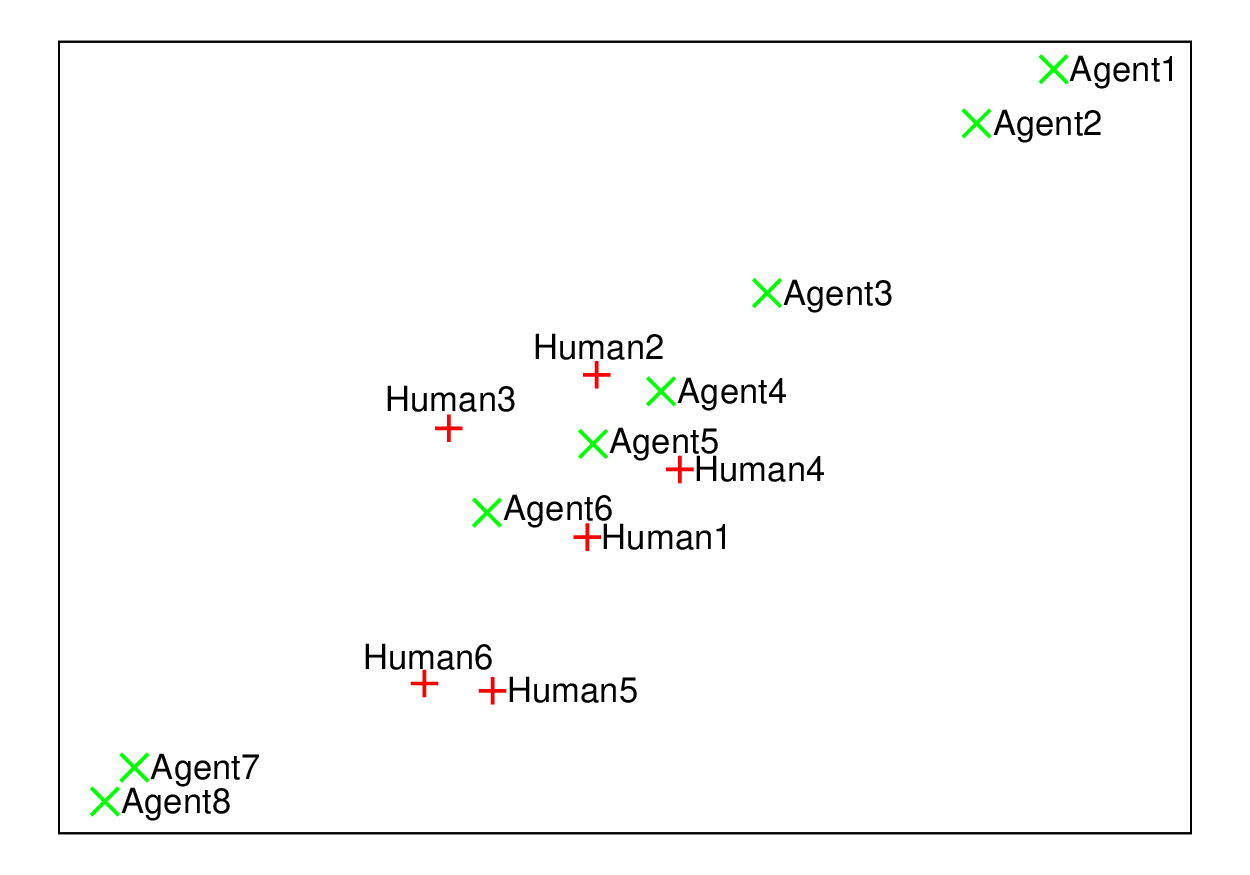}
	\caption{Result of a MDS applied to EMD matrix computed on ``velocity relative to the direction'' signatures. This representation represents well the original distances as the stress is a bit more than 0.003.}
	\label{MDSdv}
\end{figure}

\paragraph{}
In order to visualise the data, we chose to do a Principal Components Analysis (PCA) on the signatures. Subjects are represented using the two principal components in figures \ref{PCAv} and \ref{PCAdv}.
\paragraph{}
The first chart (figure \ref{PCAv}) represents very well the data with about 90\% of the variance represented. The first PCA axis is negatively correlated with brutal angle changes and the second PCA axis is positively correlated with small angle changes. The second chart (figure \ref{PCAdv}) is less accurate with only 78\% of the variance represented; however, it does suffice to extract some interesting data's features. The first PCA axis is globally negatively correlated with walking backwards and backwards+sideways. The second PCA axis is positively correlated with walking forward+slightly sideways and negatively with forward+sideways.
\paragraph{}
The most important information that both charts reveal is that there is a difference between humans' and agents' behaviours: it is possible to draw a line which separate humans from agents on the two charts. To study the influence of efficiency, agents and humans are ranked depending on their score (number of ``kills''). As humans did not play against agents, these ranks apply only within those two groups: Human1 is more efficient than Human2, which is more efficient than Human3, etc. Agent1 is more efficient than Agent2, etc. but we do not know if Agent1 is more efficient than Human1. It seems that the \emph{skill} parameter which influence the agents' efficiency have a quite artificial effect: agents are ordered on the first principal component on both charts.
\paragraph{}
Even if those results are promising, PCA suffers from a flaw in our case: in the signatures, a component $i$ is much closer in term of angle to the $i+1^{th}$ and $i-1^{th}$ than to the $i+2^{th}$ and $i-2^{th}$ component, which is not taken into account by the PCA. To calculate the distance between two signatures considering this particularity, we chose to use the Earth Mover's Distance (EMD) \cite{Rubner2000}. Figuratively speaking, the distance between two vectors $V_1$ and $V_2$ is equal to the minimum effort made to carry earth from the ``relief'' $V_1$ to the ``relief'' $V_2$. The interest of this metric is explained in the figure \ref{EMD}.
\begin{figure}[!ht]
	\includegraphics[width=\linewidth]{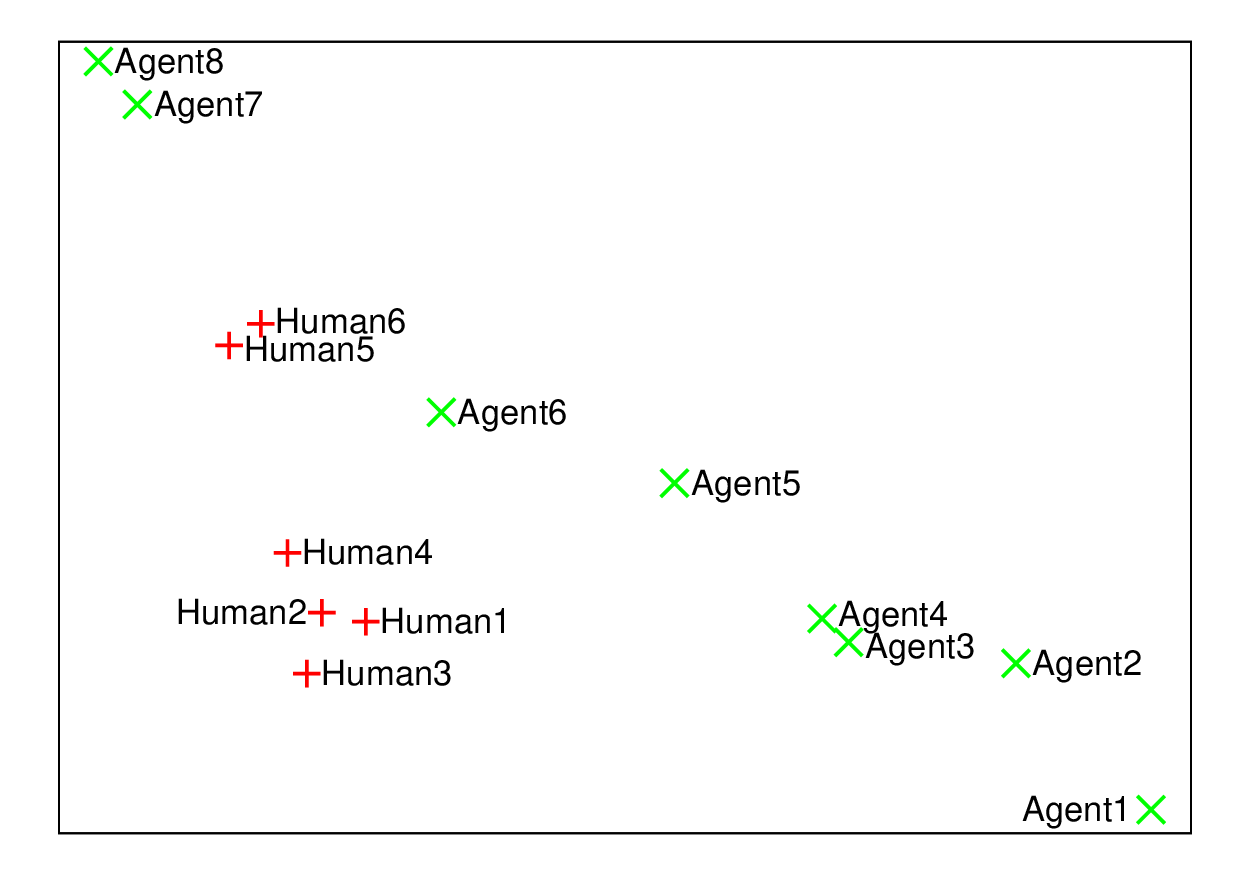}
	\caption{Result of a MDS applied to EMD matrix computed on ``velocity change angle'' signatures. This representation represents well the original distances as the stress is a bit less than 0.003.}
	\label{MDSv}
\end{figure}
\paragraph{}
With the EMD, we can build a matrix of dissimilarities by computing the distance between each signature. To visualise these dissimilarities, we used a method called MultiDimentionnal Scaling (MDS) which consists in representing each vector by a point in a plan. The goal of MDS is to approximate a given dissimilarity matrix between vectors by a matrix of Euclidean distances ($L_2$) between the points representing the vectors. The result of the MDS method applied to dissimilarities matrices computed with EMD on our signatures are given in figures \ref{MDSv} and \ref{MDSdv}. Note that there is no need for a scale because it is the relative distance between points that is important, not the real distance.
\paragraph{}
The results are a bit different from what we get with the PCA analysis: humans' signatures are much more mixed with the agents' than with the PCA. In the figure \ref{MDSdv} we can see that humans and agents have quite similar behaviours although there are still some differences between the two. We can also note that all agents are on a line in figures \ref{MDSv} and \ref{MDSdv} which reinforce the idea that the \emph{skill} parameter have a quite linear effect. It seems that this effect is not pertinent in the case of the figure \ref{MDSv} because as the humans' efficiency rise they tend to move away from the agents'.
\paragraph{}
The PCA and the MDS analysis seems to give quite different results. The MDS is based on the EMD which measure the difference between global signatures' shapes. From this point of view, agents are quite close to humans so we can think that they might be believable. However, the PCA analysis, being more strict, allows us to clearly distinguish human and artificial behaviours. This difference can come from the fact that humans have limitations to control their virtual character because they use keyboards. Evaluated agents do not copy those limitations, as a result, they may have a global human-like behaviour but they can be recognised if we look closely. Note that our two signatures are designed to study the velocity vector so all our conclusion applies only to the way agents move. We should have designed a lot more signatures in order to study much more aspects of the behaviours.

\section{CONCLUSION}
\paragraph{}
The proposed method seems promising as it could help in assessing the believability of a behaviour. Its main advantage is that it can evaluate a large number of agents, allowing finer improvement of the models' parameters. This advantage is due to the principle of signatures, vectors which characterise behaviours' aspects. We found out in our test experiment that even simple signatures can give interesting results.
\paragraph{}
However, there is still some questions that should be answered: to what extent do humans notice variations in behaviours? and do signatures have the same sensitivity as humans? Studies confronting the current method to classic ones should be done to evaluate signatures. Even if those studies are time-consuming and complex they will result in useful and reusable signatures.
\paragraph{}
The next step is to use this method in a behaviour modelling project. The goal is to optimise the models according to an evaluation method based on our present work. The final agents will then be evaluated in a classic way, with subjects judging agents in their environment.
\paragraph{}
We will, however, need to improve the type of signatures for the evaluation to be more precise. Presented signatures are \emph{global signatures}, they are computed at each time step. \emph{Contextual signatures} are computed only when certain perceptions and/or actions based conditions are met. They will provide more meaningful information about behaviours. Another new type of signature can be useful, \emph{temporal signatures} which will measure time between events. It can be used to measure reaction time which is an important factor of believability.

\bibliographystyle{myBibStyle}
\bibliography{library}

\end{document}